\documentclass{acmart}
\usepackage{multirow}
\usepackage{tabularx}
\usepackage{booktabs}

\AtBeginDocument{%
  \providecommand\BibTeX{{%
    \normalfont B\kern-0.5em{\scshape i\kern-0.25em b}\kern-0.8em\TeX}}}

\setcopyright{acmlicensed}
\copyrightyear{2018}
\acmYear{2018}
\acmDOI{XXXXXXX.XXXXXXX}

\acmConference[Conference acronym 'XX]{Make sure to enter the correct
  conference title from your rights confirmation emai}{June 03--05,
  2018}{Woodstock, NY}
\acmISBN{978-1-4503-XXXX-X/18/06}

\begin{document}

\title[Automated Academic Papers Interpretation System]{Bridging Research and Readers: A Multi-Modal Automated Academic Papers Interpretation System}

\author{Feng Jiang}
\email{jeffreyjiang@cuhk.edu.cn}
\affiliation{
  \institution{The Chinese University of Hong Kong, Shenzhen}
  \country{China}
}

\author{Kuang Wang}
\affiliation{%
  \institution{Zhejiang University}
  \country{China}}

\author{Haizhou Li}
\affiliation{%
  \institution{The Chinese University of Hong Kong, Shenzhen}
  \country{China}
}







\begin{abstract}
  In the contemporary information era, significantly accelerated by the advent of Large-scale Language Models (LLMs), the proliferation of scientific literature is reaching unprecedented levels. Researchers urgently require efficient tools for reading and summarizing academic papers, uncovering significant scientific literature, and employing diverse interpretative methodologies. To address this burgeoning demand, the role of automated scientific literature interpretation systems has become paramount. However, prevailing models, both commercial and open-source, confront notable challenges: they often overlook multimodal data, grapple with summarizing over-length texts, and lack diverse user interfaces. In response, we introduce an open-source multi-modal automated academic paper interpretation system (MMAPIS) with three-step process stages, incorporating LLMs to augment its functionality. Our system first employs the hybrid modality preprocessing and alignment module to extract plain text, and tables or figures from documents separately. It then aligns this information based on the section names they belong to, ensuring that data with identical section names are categorized under the same section. Following this, we introduce a hierarchical discourse-aware summarization method. It utilizes the extracted section names to divide the article into shorter text segments, facilitating specific summarizations both within and between sections via LLMs with specific prompts. Finally, we have designed four types of diversified user interfaces, including paper recommendation, multimodal Q\&A, audio broadcasting, and interpretation blog, which can be widely applied across various scenarios. Our qualitative and quantitative evaluations underscore the system's superiority, especially in scientific summarization, where it outperforms solutions relying solely on GPT-4. We hope our work can present an open-sourced user-centered solution that addresses the critical needs of the scientific community in our rapidly evolving digital landscape.
\end{abstract}




\maketitle

\section{Introduction}\label{sec:introduction}
In the digital information era, the rate of data production is escalating daily, a phenomenon that is also evident in the realm of scientific research. The sheer volume of scholarly papers is burgeoning at an unprecedented pace. For instance, the renowned preprint server arXiv took over 23 years to accumulate its first million submissions, yet only seven years to gather the next two million, with the subsequent million potentially arriving in just four and a half years\footnote{https://arxiv.org/stats/monthly\_submissions}. In certain scientific domains, query-based searches often yield a plethora of related articles, far exceeding human capacity for processing~\cite{altmami2022automatic}. This explosion of information is particularly amplified by the advent of Large Language Models (LLMs), which have significantly accelerated the production of scientific literature. The overwhelming abundance of research papers necessitates a transformation in how we access, comprehend, and engage with scientific knowledge, prompting the need for innovative approaches to managing and interacting with this ever-growing body of work.

\begin{figure}[!htb]
    \centering
    \includegraphics[width=\textwidth]{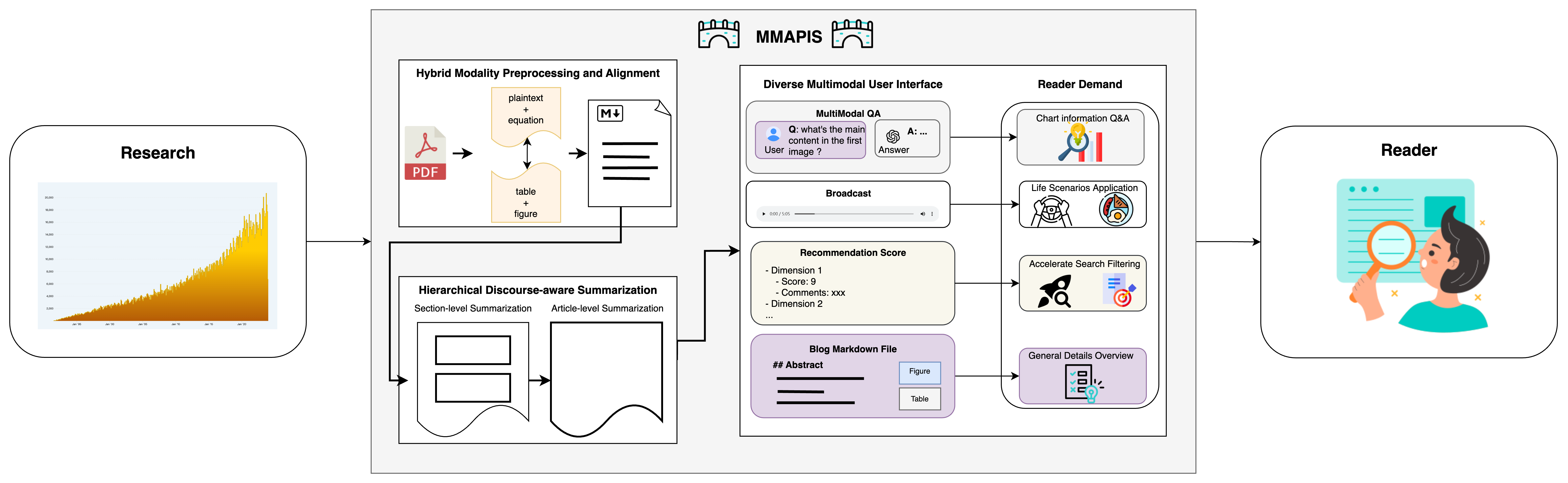}
    \caption{MMAPIS: Bridging Research and Readers. }
    \label{Bridging}
\end{figure}

Recent trends in research have seen the gradual emergence of paper interpretation systems, with many commercialized service platforms~\footnote{https://papers.cool/ \\ https://www.paperdigest.org/ \\ https://www.paperreading.club/ } progressively coming online. These systems are capable of swiftly extracting the gist of entire papers from academic websites like arXiv and providing interpretations in the form of summaries. The latest GPT-4~\footnote{https://chat.openai.com/?model=gpt-4} model also offers the capability to upload PDF documents and interpret papers by designing specific prompts. However, their proprietary nature renders their operational processes opaque and raises concerns about information security when uploading papers to third-party websites. On the other hand, some open-source works~\footnote{https://github.com/Anil-matcha/ChatPDF \\
https://github.com/kaixindelele/ChatPaper} focus on converting PDF documents into text using OCR (Optical Character Recognition) technology, followed by feeding the text into summarization models or Large Language Models (LLMs) to obtain the final interpretation. Despite the advancements in paper interpretation systems, fulfilling the aforementioned functional requirements presents several challenges:


\textbf{Overlooking Multimodal Data in Academic Papers}: Most existing paper interpretation systems primarily treat papers as the text for summarization, overlooking the structure information and other modalities, such as mathematical formulas, tables, figures, and so forth, that encapsulate the most crucial experimental results, concepts, or workflows~\cite{bhatia2012summarizing}. This singular focus on textual information fails to capture the full richness of multimodal data, which is often crucial in scholarly papers.

\textbf{Grappling with summarization for over-length text}: Previous methods of summarization have been limited in their ability to handle longer texts. Although current large-scale language models (e.g., those capable of processing texts up to 100K words~\cite{chen2023longlora}) can manage over-length texts, they often struggle to capture complex details~\cite{liu2023lost}. On the other hand, segmented summarization approaches typically employ length truncation or fixed grouping methods, leading to semantic incompleteness. Additionally, each section of a paper has different focal points, necessitating distinct considerations for their summarization.

\textbf{Lacking Diverse User Interface}: The existing systems largely overlook the potential of multimodal and varied downstream task scenarios. They typically only display results as summaries or engage in text-based chats about the paper. This limitation can inconvenience many users, especially those in related fields who would benefit from more diverse methods of paper interpretation.

To address the aforementioned challenges, we introduce an open-source, multimodal academic paper interpretation system (MMAPIS)~\footnote{Our code will be released at \url{https://github.com/fjiangAI/MMAPIS}.} bridging research and readers, as shown in Figure~\ref{Bridging}. It comprises three key phases: multimodal PDF information processing and alignment, hierarchical discourse-aware summarization, and a variety of multimodal user interfaces. Our system adeptly processes scientific literature in PDF format from sources like arXiv or user-uploaded documents, extracting structured text, images, tables, and equations separately to consider multimodal information comprehensively. Then, by employing a discourse-aware approach for sectional summarization, the system ensures that all critical information within a paper is captured, and there will be different focuses for each specific section, especially the key parts, such as methods and models that often appear in the middle of the document. Finally, we offer a diverse array of multimodal downstream user interfaces, including paper recommendations, multimodal Q\&A, audio broadcasting, and interpretation blogs. 



Specifically, our system begins by separately processing the text and other modalities (images, tables) of PDF-format papers. For the text, we employ the advanced Nougat model to convert it into a markdown format with rich text, which includes structured markup to indicate the structure of the document and formulas. For other modalities, we use PDFFigures to extract images and tables contained within each section of the paper. Then, we align information from different modalities using section names, allowing content from various modalities to be attributed to the same section, which facilitates the use in subsequent applications. Next, we introduce a hierarchical, discourse-aware summarization method to effectively alleviate the process of textual constraints imposed by most summarization methods as they are often limited in max tokens, due to memory complexities, hardware constraints, and time-consuming pretraining processes~\cite{devlin2018bert,zhang2020pegasus}. Initially, we use the extracted section boundaries to divide the paper into several parts. Each part is then summarized using a Large Language Model (LLM) with special prompts. Subsequently, these individual summaries are consolidated based on requirements, ensuring that important information from each section of the document is retained and not overlooked. Finally, we have developed four common types of interactive applications by utilizing LLMs with various tools and specifically designed prompts. These include paper recommendation applications that score the quality of papers, detailed blog interpretations of paper content, more convenient audio readings, and multimodal interactive Q\&A. Our system also supports the customization of downstream applications by providing the necessary APIs. Our contributions are summarized by the following:

\begin{itemize}
\item  We propose an open-source multimodal paper interpretation system that provides a clear and concise solution to enhance readers' understanding and efficiency of scientific papers. It can effectively process scientific papers from sources such as arXiv or user uploaded PDFs, generating different forms of paper interpretation.

\item  We introduce a multimodal alignment method to preprocess PDF-format scientific papers, separately handling text and visual elements (images and tables) followed by a section-wise alignment process. It ensures coherent integration of text, images, and tables within the same section, enhancing the comprehension of multimodal information in scientific papers.

\item  We develop a hierarchical, discourse-aware summarization method for dealing with long scientific documents. It utilizes structured text extracted from documents to produce concise summaries at both section and document levels, ensuring the retention of essential information.

\item  We offer a range of user interfaces to present paper interpretation results. It provides users with insights in various formats including paper recommendations, multimodal Q\&A, audio broadcasting, and interpretation blogs, enhancing user engagement with scientific papers in various scenarios.
\end{itemize}

\section{related work}
\textbf{Dealing with Source Paper in Academic Papers Interpretation Systems}. 
The efficiency of an interpretation system hinges largely on the quality of data retrieved from PDF files. Historical reliance was placed on Optical Character Recognition (OCR) engines to distill plaintext for subsequent processing in earlier interpretation systems. While these engines have shown efficacy in extracting individual characters and words from images, their line-by-line approach fails to preserve relative positional relationships among different formats, particularly with regard to mathematical expressions and tables~\cite{blecher2023nougat}. Moreover, multimodal elements embedded within the documents often elude these OCR engines. Present trends in interpretation systems favor the utilization of efficient PDF Parsing Libraries, exemplified by systems such as D2S\cite{sun2021d2s} and ChatPaper~\footnote{https://github.com/kaixindelele/ChatPaper}, which utilize Grobid, and ChatPDF~\footnote{https://github.com/Anil-matcha/ChatPDF}, based on PyPDFium2Loader. Relative to OCRs, these libraries excel in analyzing PDF document structure to extract richer metadata, including embedded images. However, they still face challenges in retrieving location-specific information, such as formulas, and exhibit limitations when dealing with scientific documents that integrate images directly into the PDF format, hence, their utility remains somewhat limited. Their inherent object-based processing approach, coupled with a tendency to overlook the integrity of structural discourse, results in an inability to semantically connect different objects, necessitating the deployment of an additional Information Retrieval (IR) model for alignment ~\cite{sun2021d2s}.

 \textbf{Summarization in Academic Papers Interpretation Systems}. As a crucial component of paper interpretation systems, extracting key information from academic papers is often approached as a summarization task. Since the 1950s, the field of generic text summarization has seen significant progress~\cite{el2021automatic}. However, summarizing academic papers presents unique challenges, given their structured nature with typical sections such as the introduction, methodology, experiments, and conclusions. The excessively long text poses a challenge for directly summarizing~\cite{teufel2002summarizing}. Previous studies have indicated that despite the ability of large-scale language models to process inputs of up to 100,000 words~\cite{chen2023longlora}, they tend to disproportionately lose information from the middle sections of texts~\cite{liu2023lost}. Reducing the text to shorter text blocks is the mainstream approach. On the one hand, the methods employed include selecting key sentences or words~\cite{yang2016amplifying,slamet2018automated,hetami2015perancangan} and using segmented sliding windows~\cite{raffel2020exploring} in conjunction with attention mechanisms~\cite{beltagy2020longformer,zaheer2020big}. They do not consider the inherent structured information of the text, which can easily lead to incomplete semantic segmentation of the obtained abstract. On the other hand, the abstracts of different parts have different focuses, such as the abstract and methodology sections, and a unified abstract cannot meet their characteristics.

\textbf{User Interaction in Academic Papers Interpretation Systems}. Current paper interpretation systems primarily employ two types of user interaction models: textual summaries and dialog-based interpretations. Textual Summary Systems: These systems generate concise text-based summaries of scientific papers\footnote{https://papers.cool/,https://hub.baai.ac.cn/papers}. While efficient in providing quick overviews, they often lack depth and neglect the multimodal aspects of papers, such as graphs and tables. Dialog-based Interpretation Systems: These systems engage users in an interactive dialogue format~\footnote{https://github.com/arc53/DocsGPT,https://chat2doc.cn/,}, allowing for query-based information retrieval. However, their adaptability is limited, particularly in handling multimodal content (such as figures and tables) and varying interaction scenarios like blog interpretation or audio readings. While these systems offer basic interpretative insights, they commonly fall short of fully supporting multimodal content and lack flexibility across different application scenarios. 



\section{The framework of Multi-Modal Automated Scientific Paper Interpretation System}
\begin{figure}[!htb]
    \centering
    \includegraphics[width=\textwidth]{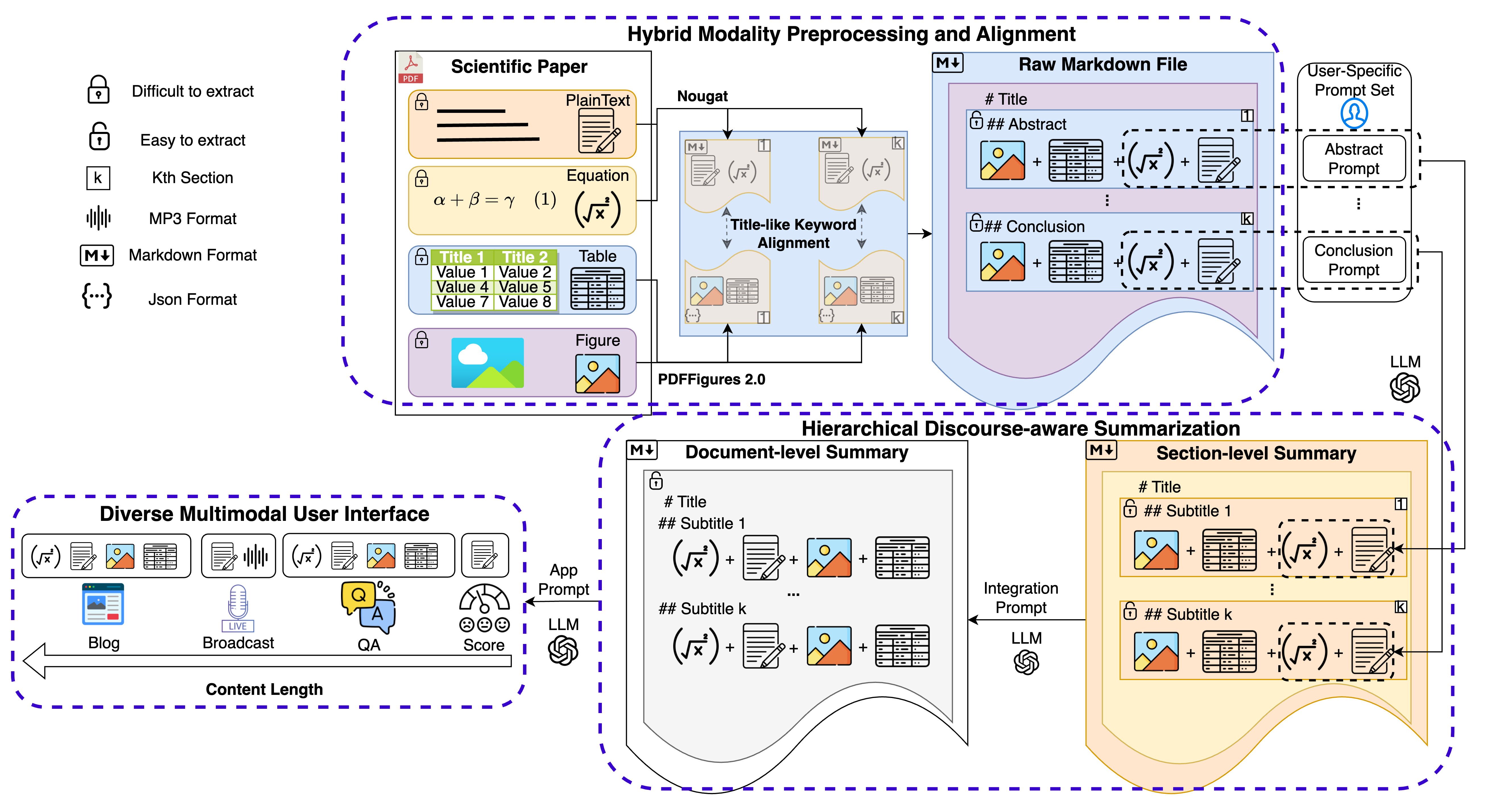}
    \caption{The framework of the multi-modal automated academic paper interpretation system (MMAPIS). }
    \label{workflow}
\end{figure}

As highlighted in Section~\ref{sec:introduction}, paper interpretation systems are crucial for aiding academic research, rapidly extracting key information, and enhancing the efficiency of literature review processes. However, current systems exhibit notable limitations in handling multimodal data (such as text, images, and equations), processing lengthy documents, and providing diverse user interfaces. To address these challenges, we have developed an open-sourced multi-modal automated academic paper interpretation system (MMAPIS), whose design and components are depicted in the accompanying Fig.~\ref{workflow}. Our system comprises three main parts: (1) \textbf{Hybrid Modality Preprocessing and Alignment Module}; (2) \textbf{Hierarchical Discourse-Aware Summarization Module}; (3) \textbf{Diverse Multimodal User Interface Module}. Firstly, the Hybrid Modality Preprocessing and Alignment Module effectively processes and integrates different types of information from papers separately, including text, images, and tables. Next, the Hierarchical Discourse-Aware Summarization Module utilizes advanced LLMs with special prompts to extract key information from each section of a paper, generating comprehensive and accurate summaries. Lastly, our system features a Diverse Multimodal User Interface Module, designed to meet the specific needs of various user groups, offering multiple interaction modes like paper recommendations, multimodal Q\&A, audio broadcasting, and interpretation blogs.



\subsection{Hybrid Modality Preprocessing and Alignment Module}
\label{sec:Multimedia Alignment}
Given the difficult task of manipulating multimodal content present in PDFs, our objective is to reconstruct the source document into semantically similar counterparts, such as in Markdown format, following Hybrid Modality Preprocessing. The reformed document can facilitate subsequent alignment modules and produce semantically comprehensive, high-quality summaries by preserving rich multimodal information (such as tables and figures) and the hierarchical discourse structure (such as sections).

Unlike traditional methodologies predominantly relying on OCR engines or efficient libraries, which fail to discern position-sensitive modal details and ignore document structure specifics, our preprocessing approach is inspired by Nougat~\cite{blecher2023nougat} and PDFFigures 2.0~\cite{clark2016pdffigures} to respectively extract text and other modalities. Nougat provides an end-to-end trainable encoder-decoder transformer-based model, adept at predicting text information in pictures, including plaintext and mathematical formulas in the screenshots of each PDF page, to markdown format. We utilize Nougat as a tool for extracting text and its inherent hierarchical structure between different paragraphs from PDFs, minimizing data loss during conversion and thereby significantly reducing the complexity of the subsequent segmentation. In addition, PDFFigures 2.0~\cite{clark2016pdffigures}, a methodology that identifies figures and tables by reasoning about the empty regions within the text, provides not only the images but also their attribution information. This reduces the difficulty of multimodal alignment when assigning affiliations across modalities based on their relative structural information to layout tokens, such as section titles.

The results of both methods, Nougat and PDFFigures 2.0, complement each other—Nougat yields plaintext and mathematical formulas, whilst PDFFigures 2.0 provides screenshots of figures and tables.  Notably, these components extracted plaintext or figures with the section name, which can be intuitively aligned with their corresponding title-like keywords, thereby facilitating the recreation of parsing results that closely adhere to the structure of the source document. Specifically, all tables and figures parsed from the PDF are retained after aligning the sections they belong to due to their value in encapsulating the most critical experimental results and concepts in scientific documents~\cite{bhatia2012summarizing}.

\subsection{Hierarchical Discourse-aware Summarization Module}
To address the challenges of long documents frequently encountered in interpretation systems, we introduce a two-stage summary process, Hierarchical Discourse-aware Summarization, to generate a holistic interpretation.  This methodology could alleviate the limitations of previous works (such as semantic fragmentation and efficiency detriments) by splitting documents with section names given by the first module and generating the summary within and between sections via LLMs.


The first stage of our methodology is devoted to the formation of a section-level summary. The process first involves partitioning the document into sections using the hierarchical cues embedded in the title within the Markdown. This approach diverges from the predefined four sections proposed by FacetSum~\cite{meng2021bringing}, which might only be universally applicable to some papers. Instead, our approach favors dissection at almost every individual section. This flexibility permits a reduction in the lengths of processing segments whilst ensuring the maintenance of section-level and document-level semantic integrity without overlap. Further, it offers adaptive control to expand or contract sections as needed, mitigating noise interference. For example, sections such as the appendix and references can be excised, promoting a more streamlined content-processing experience. Segments such as abstract, introduction, etc, are then matched with corresponding prompts from our prearranged set, specifically designed to facilitate the extraction of key points. This step operates under the assumption that scientific documents adhere to a generic structure~\cite{gehrmann2018bottom}, and the primary function of each section remains relatively constant. The user-specific prompt encourages the LLMs with the dependence that aligns with the reader's needs and expectancy~\cite{he2020ctrlsum,wu2021controllable} for multi-faceted insight. For sections whose titles are absent from the prompt set, we assign a universal prompt as a viable alternative. 

Upon generating section-level summaries, we proceed to a prompt-guided integration stage to construct a document-level summary. It mainly focuses on fortifying the cohesion and continuity between sections, thereby guiding the downstream application. Notably, during the integration stage, we include the title, authors, and affiliation information filtered through NER technology within the reference text, enhancing the synthesized summary's affinity.

\subsection{Diverse Multimodal User Interface Module}


Our interpretation system, enhanced by a user-friendly Streamlit-based interface, adeptly transforms the outputs from the Hierarchical Discourse-Aware Summarization Module into four distinct downstream applications, each tailored to fit different user scenarios and ordered by the generated content-length:

1. \textbf{Paper Recommendation}: As the first application, our system employs meticulously designed LLM prompts to evaluate papers across five critical dimensions: clarity of objectives and central themes, appropriateness and accuracy of methods, authenticity and precision of data and findings, depth and conclusiveness of analysis, and overall writing quality. The primary four document-level indicators rely on generated summaries for token reduction assessment. However, evaluating overall writing quality, a more fine-grained metric at the paragraph or even word level, necessitates the use of the original text for a fair assessment. Due to the potential for token overflow and assumptions about writing consistency, considering the reader's inclination to first read the beginning and end, the evaluation is applied to strategically selected excerpts, specifically the introductory and concluding sections. These features aim to swiftly gauge the quality of a paper, providing users with immediate insights into its merits.

2. \textbf{Multimodal Q\&A}: Advancing beyond the confines of conventional text-based question-and-answer formats, our system introduces an amplified two-tier Q\&A  feature that incorporates specific queries about figures or tables extracted from the papers. With the user's request for an in-depth elucidation of a graphic representation within the paper, we initially employ GPT-3.5 Turbo to discern the chart's provenance from user's questions, that is, the index of the illustration juxtaposed with the section to which it pertains, e.g. ("Introduction",1) to locate the target picture. Then we feed user queries and relevant images to GPT-4 to deliver a comprehensive interpretation. This functionality leverages GPT-4's multimodal processing abilities, enabling more precise and targeted responses, thereby enriching the user's understanding of the paper's content.

3. \textbf{Audio Broadcasting}: Recognizing the need for quick assimilation of information in real-time scenarios, our system introduces a feature tailored for the generation of colloquial broadcast scripts. Utilizing prompts with ChatGPT for the formulation of straightforward sentences, it synthesizes narratives suitable for verbal dissemination predicated on the produced summary. Then audio broadcasting is generated through text-to-speech (TTS) interfaces, such as Azure TTS~\footnote{https://azure.microsoft.com/en-us/products/ai-services/text-to-speech/} or Youdao TTS, thereby providing users with a distinctive and user-friendly method to engage with the research.

4. \textbf{Interpretation Blog}:  The system provides an interpretation blog feature that demands detailed and thorough insights for a more in-depth exploration of the paper. This tool leverages LLM-grounded prompts to render interpretative blogs derived from the created summary, fostering an extensive comprehension of the paper's central subject matter and intricate technical elements through aligning Title-like Keywords to integrate other modalities. Special emphasis is placed on readability and adherence to established blog formats, ensuring a seamless narrative flow from a clear introduction to a conclusive ending.


\subsection{Prompt Design}

Since Transformer-based models are better optimized towards short document language tasks rather than long documents~\cite{liu2023lost}, our goal is to ensure the prompt of each role preserves unique responsibilities while maintaining brevity. Ideally, crucial information should be located either at the onset or the end of input text, and the prompt should be kept brief to optimize performance. We thus categorize prompts into three distinct segments and more details are shown in Appendix:
\begin{itemize}
    \item  \textbf{Task Description}: This segment, the prompt for the "system" role, gives an overview of the task, describing the needs, objectives, or background and stipulates the desired format of the output.
    
    \item \textbf{Current Input}: The "user" role inputs text from the raw document content, previously generated summary, or both, serving as a source of external knowledge to better comprehend the entire text.
    
    \item \textbf{Output Indicator}: This segment for the "system" role defines a specific workflow, providing guidelines for GPT to follow.

\end{itemize}

In the realm of output indicators, We adroitly blend the technologies of Chain of thought (CoT) and Chain of Density (CoD)~\cite{adams2023sparse},  each selected for its unique benefits. CoT stands out for its robustness and ability to significantly surpass the standard baseline~\cite{wei2022chain}, CoD,  in turn, expertly manages information density to harmonize between informativeness and intelligibility. 

For generating section summaries or downstream applications, e.g., broadcasts and blogs, we employ CoT as a tool for introspective evaluation and continuous enhancement. Particularly when producing section summaries, we blend the task description and output indicator to ensure task clarity and condense output requirements while stipulating the specific output format in downstream applications, considering the varying requirements of diverse output modes.

Conversely,  in the process of merging section summaries or regenerating, we apply CoD to mitigate any potential loss of valuable entities while simultaneously boosting information density, thereby providing a more comprehensive and informative interpretation.

\section{Demo \& Evaluation}

\subsection{Demo}
\label{sec:demo}

\begin{figure}[!htb]
    \centering
    \includegraphics[width=\textwidth]{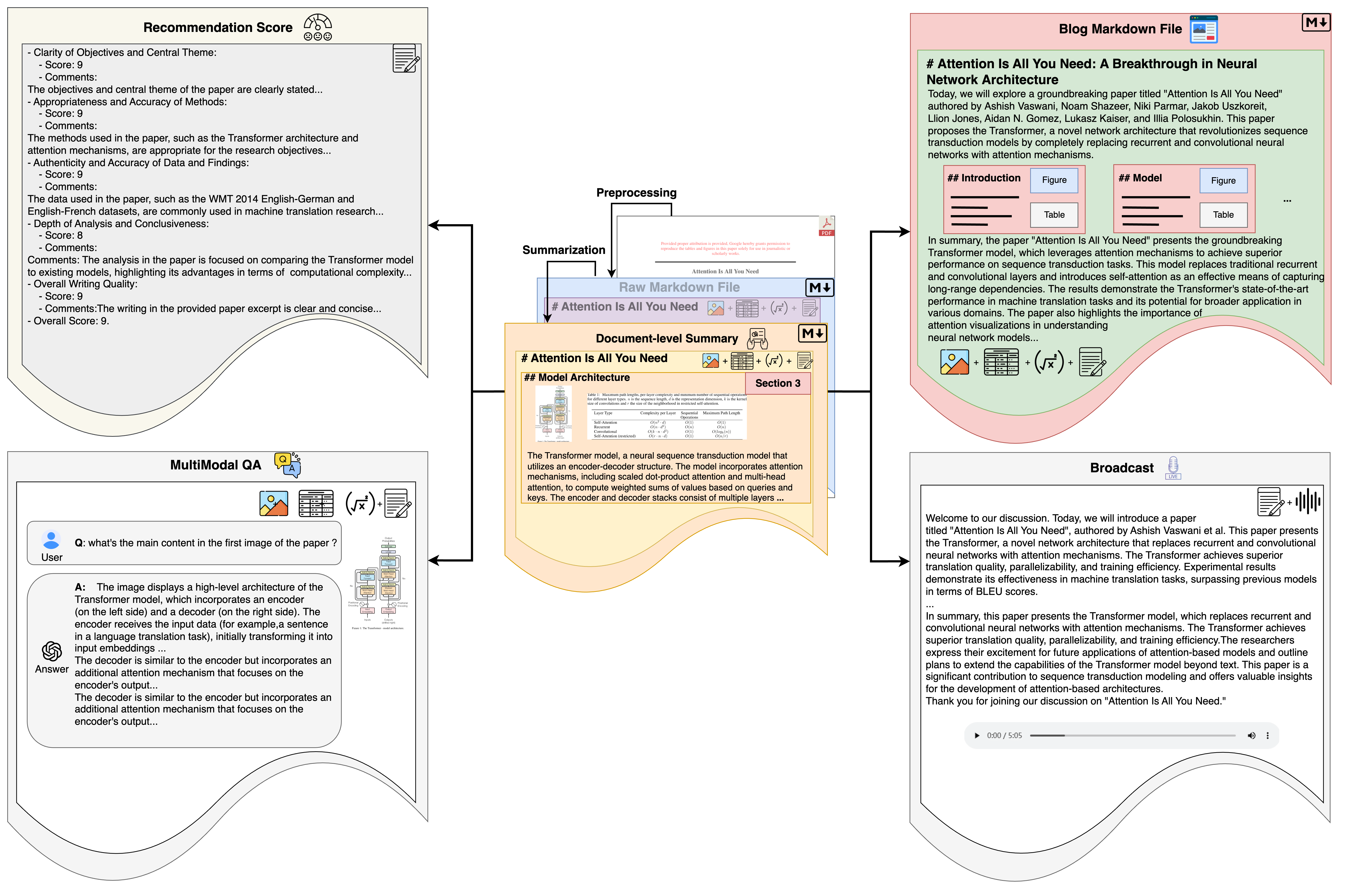}
    \caption{The demos of the diversified user interfaces in MMAPIS (Optimized for enhanced visual presentation).}
    \label{demo}
\end{figure}

In Fig.~\ref{demo}, we showcase our versatile and adaptable downstream applications, each tailored for different scenarios. The case study revolves around the seminal paper titled ``Attention is All You Need''~\cite{vaswani2017attention}, which introduced the revolutionary ``transformer'', aimed at facilitating observations. Initially, we present a recommendation score in markdown format anchored in the top left corner, enabling academics to promptly assess the quality of papers across five distinct dimensions. Subsequently, our multimodal Q\&A mechanism enhances interpretive comprehension through a dialogue-oriented exploration of key content, including data encapsulated in figures, as exemplified by the lower left corner. To address scenarios that demand convenience, such as engaging with content while driving or during mealtimes, we provide an audio feature in MP3 format, offering a preliminary understanding of the manuscript's narrative. Ultimately, the top right corner displays the resulting blog posts in markdown format, retaining almost all metadata extracted from the source document, including figures, tables, mathematical formulas, and plain text, emphasizing readability and audience engagement, thereby expediting comprehension of the paper's core subject.

\subsection{Quantitative Analysis for Summarization}

In addition to the case studies mentioned above, we conducted a quantitative analysis focusing on the core aspect of paper interpretation systems – the quality of paper summarization. Specifically, we first selected 100 papers from arXiv to form our test set. Since the training data for GPT-4 ends in April 2023, to avoid performance bias due to data leakage, we sampled 50 papers each from December 2017 and December 2023 on arXiv, forming two test subsets named CS2017 Dataset and CS2023 Dataset, respectively. The statistical information of these subsets is shown in Table ~\ref{csdatainfo}. For our baseline system, we tested the currently best-performing model, GPT-4, to compare its performance against our model. In terms of evaluation metrics, we adopted the COD approach, which includes five dimensions: Informative, Quality, Coherence, Attributable, and Overall~\cite{adams2023sparse}. It involves using GPT-4 to score the summaries instead of using ROUGE scores due to the absence of reference texts and the inability of ROUGE scores to identify overlaps between synonymous tokens or phrases~\cite{bhandari2020re,chaganty2018price,hashimoto2019unifying,kryscinski2019neural}.



\begin{table}[!htb]
    \caption{Statistical information of CS datasets for 2023 and 2017.}
    \label{csdatainfo}
    \centering
    \begin{tabular}{lccc}
        \hline
        Dataset & Ave. Section Length (tokens) & Ave. Document Length (tokens) & Ave. Number of Sections \\ \hline
        CS2023 & 1364 & 9152 & 6.71 \\ 
        CS2017 & 1163 & 7730 & 6.64 \\ \hline
    \end{tabular}
\end{table}

\subsubsection{Result \& Analysis}

\begin{table}[!htb]
    \caption{The performance of summary generation. Each sample is evaluated three times: the main value is the overall average score, and the subscript is the overall standard deviation. }
    \label{res}
    \centering
    \small
    
    \begin{tabular}{lccccccc}
    \hline
    Dataset & Summarizer   & Informative & Quality & Coherence & Attributable & Overall & Eval Average \\ \hline
    \multirow{2}{*}{CS2017} 
        & Ours  & \textbf{$4.534_{0.256}$} & \textbf{$4.440_{0.297}$} & \textbf{$4.518_{0.349}$}  & \textbf{$4.568_{0.381}$} & \textbf{$4.521_{0.240}$} & \textbf{$4.516_{0.305}$} \\ \cline{2-8}
        & GPT-4  & $4.392_{0.246}$ & $4.350_{0.282}$ & $4.444_{0.330}$ & $4.554_{0.324}$ & $4.434_{0.200}$ &  $4.435_{0.276}$ \\ \hline
                            
    \multirow{2}{*}{CS2023} 
        & Ours  & $4.498_{0.204}$ & $4.376_{0.344}$ & $4.455_{0.395}$ & $4.439_{0.516}$ & $4.454_{0.250}$ & $4.444_{0.342}$ \\ \cline{2-8}
        & GPT-4  & $4.363_{0.260}$ & $4.317_{0.268}$ & $4.429_{0.263}$  & $4.460_{0.462}$ &  $4.377_{0.232}$ & $4.389_{0.297}$ \\ \hline

    \end{tabular}
\end{table}

As demonstrated by the results, our system surpasses GPT-4 in nearly all aspects in both the CS2023 Dataset and CS2017 Dataset through Horizontal comparison. Specifically, in the dimension of Informative and Overall, our methodology exhibits a remarkable advancement compared to the generalized summary offered by GPT-4 and remains relatively stable, presenting that the result of MMAPIS offers denser entity density and detailed information of key specific, which tends to be more favored by humans, as demonstrated in ~\cite{adams2023sparse} that the overall dimension has the highest summary-level Pearson Correlation to human preference, while others also maintain a positive correlation ranging from 0.120 to 0.245. Such evidence underscores the efficient performance of  Hierarchical Discourse-Aware Summarization, which reduces the likelihood of information loss during segmentation and enables the summarizer to yield a comparatively detailed summary, thereby accurately encapsulating the key narratives and catering to human preference.

The underlying rationale for this observation is that the Hierarchical Discourse Summarization mitigates the 'layout bias' identified by Kryściński et al. ~\cite{kryscinski2019neural}, i.e., approximately 60 \% of essential sentences are located within the opening 30\%. Unlike conventional methods that favor extraction or truncation based on empirical knowledge or model design, our method veers away from layout bias by using data that closely resembles the original text, which preserves the semantic and structural integrity of the information, even when filtering out particular sections that are less valuable to the readers. It's supported by ~\cite{koh2022empirical}, which showed that salient content is more uniformly dispersed throughout long documents, in contrast to models that often benefit from layout biases in short documents ~\cite{gehrmann2018bottom,paulus2017deep,see2017get}. 

The notable performance enhancement may also be attributed to the interesting observation connected with the hypothesize of the "lost in the middle" phenomenon ~\cite{liu2023lost} that LLMs exhibit a `U-shaped' performance curve with a strong reliance on information appearing at the beginning and end. Specifically, as our method segments the text based on the document's structure, the average length of each section within the documents of CS2017 Dataset and CS2023 Dataset is maintained below 2000 tokens, specifically averaging at 1163 and 1364 tokens, respectively. This approach substantially mitigates any potential performance degradation instigated by the position sensitivity in comparison to processing the entire text body, which encapsulates approximately 7,730 to 9,152 tokens, respectively, representing nearly a 6.7-fold increase in length. As evidenced by ~\cite{liu2023lost}, where GPT-3.5 Turbo's QA performance can drop by over 20\% with approximately 4000 token inputs while maintaining considerable performance with around 2000 token inputs. It can also be inferred that the minimum grid of Hierarchical Discourse-Aware Summarization, which is section-based in the first stage, represents a trade-off between speed and accuracy.

In a longitudinal comparison, both GPT-4 and the MMAPIS exhibit a discernible performance deterioration across all dimensions. Interestingly,  in document-free dimensions,e.g., Quality and Coherence, the degradation is less severe, while in dimensions related to technical intricacies and knowledge reserves,  observe a more marked reduction, potentially owing to the reoccurring issue of `hallucinations' - a common trait amongst GPT models. Another contributing factor could be the average length increase in data from the CS2017 Dataset, i.e., 1163, to that of CS2023 Dataset, i.e.1364, raising the computational load and subsequently making the GPT model more prone to distractions.

\section{Conclusions and Future Work}

In this paper, we introduced an open-source Multi-Modal Automated Academic Paper Interpretation System (MMAPIS), comprising three integrated modules: (1) Hybrid Modality Preprocessing and Alignment Module; (2) Hierarchical Discourse-Aware Summarization Module; (3) Diverse Multimodal User Interface Module. Compared with existing works, our system adeptly harnesss the multimodal information in academic papers through hybrid modality preprocessing. Moreover, the hierarchical discourse-aware summarization method via LLMs with special prompts ensures that essential information across various sections of lengthy scientific texts is accurately captured and retained. Additionally, the system's diverse range of multimodal user interfaces enhances the accessibility and utility for both readers and developers. We demonstrate the efficacy and superiority of our system through qualitative demonstrations and quantitative analyses. In the future, we aim to further augment our system's capabilities, focusing on advanced integrations and optimizations. While our current design already addresses key challenges in academic paper interpretation, we recognize the potential for incorporating a broader spectrum of external knowledge and inter-document connections. This enhancement will facilitate a more nuanced understanding of academic content, especially in relation to user-specific contexts and profiles. Additionally, we are dedicated to refining the efficiency and responsiveness of our system. Our goal is to transition from primarily offline processing to more dynamic, real-time interpretations, thereby broadening the system's applicability.

\bibliographystyle{ACM-Reference-Format}
\bibliography{sample-base}

\section*{Appendix}
\label{sec:appendix}
\begin{figure}[!htb]
    \centering
    \includegraphics[width=\textwidth]{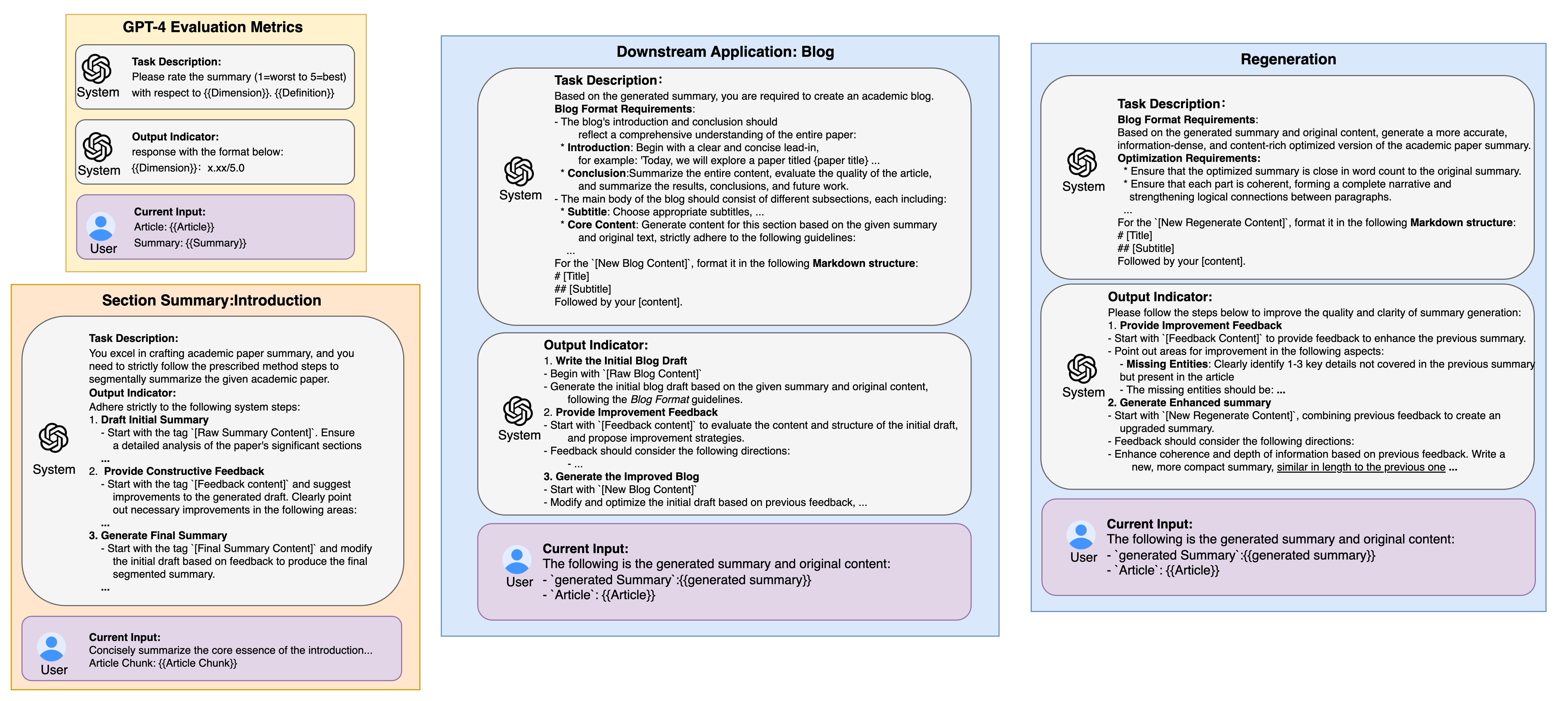}
    \caption{Prompt Example.}
    \label{prompt}
\end{figure}

In Figure \ref{prompt}, we present various examples of thoughtfully designed prompts. These include a quantifiable evaluation with GPT-4 situated at the upper left, a summary section signified by an introduction exemplar at the lower left, an illustration of application generation, specifically blog interpretation, in the central portion, while showcasing regeneration leveraging CoD technology towards the right.

The process of generating section summaries or applications for downstream use is steered by the CoT, which directs the introspective evaluation and ongoing refinement of the GPT. This process consists of three distinct steps: (1) Draft generation, (2) Self-review based on pre-determined parameters, and (3) Refinement to generate the final outcome. The main emphasis in this procedure is on fluency, authenticity, and integrity.

On the other hand, at the integration stage or regeneration, the CoD is implemented in an identical workflow. However, the principal aim here is to limit the loss of entity-specific information. During the evaluation phase, we have adopted the methods discussed by \citet{adams2023sparse} , using GPT-4 as proxy to rate performance across five dimensions: Informativeness, Quality, Coherence, Attribution, and Overall Impact.

\end{document}